%% file: acl_latex.tex
\pdfoutput=1

\documentclass[11pt]{article}

\usepackage[]{acl}

\usepackage{times}
\usepackage{latexsym}
\usepackage{fdsymbol}

\usepackage{xcolor}
\usepackage{colortbl}
\usepackage{hhline}
\usepackage{CJKutf8}

\usepackage[T1]{fontenc}

\usepackage[utf8]{inputenc}
\usepackage{xspace}

\usepackage{microtype}
\usepackage{subcaption}
\usepackage{arydshln}

\newcommand{\miast}{MIA shared task}
\newcommand{\xorqa}{XOR-TyDi QA\xspace}
\newcommand{\mkqa}{MKQA}
\newcommand{\Sref}[1]{Section~\ref{#1}}

\definecolor{amber}{rgb}{1.0, 0.75, 0.0}

\newcommand{\Tref}[1]{Table~\ref{#1}}
\newcommand{\Fref}[1]{Figure~\ref{#1}}

\usepackage{booktabs}
\usepackage{graphicx}

\title{MIA 2022 Shared Task: Evaluating Cross-lingual Open-Retrieval \\
Question Answering for 16 Diverse Languages}

\author{
  \parbox{0.9\linewidth}{\centering Akari Asai$^{\clubsuit}$, Shayne Longpre$^\spadesuit$, Jungo Kasai$^{\clubsuit}$, Chia-Hsuan Lee$^{\clubsuit}$, \\ Rui Zhang$^\vardiamondsuit$, Junjie Hu$^\heartsuit$, Ikuya Yamada$^{\bigstar\lozenge}$, Jonathan H. Clark$^{\ddagger}$, Eunsol Choi$^\dagger$} \\
 $^\clubsuit$University of Washington~~$^\spadesuit$Massachusetts Institute of Technology~~$^\vardiamondsuit$Penn State University \\ 
 $^\heartsuit$University of Wisconsin-Madison~~$^{\bigstar}$Studio Ousia~~$^\lozenge$RIKEN \\
 $^\ddagger$Google Research~~$^\dagger$The University of Texas at Austin\\
  \texttt{mia.nlp.workshop@gmail.com}  \\
}

\begin{document}

\maketitle

\begin{abstract}
We present the results of the Workshop on Multilingual Information Access (MIA) 2022 Shared Task, evaluating cross-lingual open-retrieval question answering (QA) systems in 16 typologically diverse languages.  
In this task, we adapted two large-scale cross-lingual open-retrieval QA datasets in 14 typologically diverse languages, and newly annotated open-retrieval QA data in 2 underrepresented languages: Tagalog and Tamil. Four teams submitted their systems. 
The best system leveraging iteratively mined diverse negative examples and larger pretrained models achieves 32.2 F1, outperforming our baseline by 4.5 points. 
The second best system  uses entity-aware contextualized representations for document retrieval, and achieves significant improvements in Tamil (20.8 F1), whereas most of the other systems yield nearly zero scores. 
The official leaderboard\footnote{\url{https://eval.ai/web/challenges/challenge-page/1638/leaderboard}} and baseline models\footnote{\url{https://github.com/mia-workshop/MIA-Shared-Task-2022}} are publicly available.
\end{abstract}

\input{sections/intro}
\input{sections/task}
\input{sections/data}
\input{sections/baselines}
\input{sections/submissions}
\input{sections/results}

\input{sections/analysis}
\input{sections/conclusion}

\section*{Acknowledgements}

We would like to acknowledgments and Noah A. Smith for serving as our steering committee. 
We are grateful to Google for providing funding for our workshop. 
We thank GENGO translators to translate questions into Tamil and Tagalog. 
we thank the EvalAI team, particularly Ram Ramrakhya, for their help with hosting the shared task submission site. 
We thank Maraim Masoud for her help in error analysis. 
\bibliography{custom}
\bibliographystyle{acl_natbib}

\end{document}

%% file: sections/intro.tex
\section{Introduction}
\label{sec:intro}
Open-retrieval\footnote{Also sometimes referred to as \textit{open-domain} QA; we use \textit{open-retrieval} as it is not ambiguous with the sense of ``covering many domains.''} question answering (QA) is a task of answering questions in diverse domains given large-scale document collections such as Wikipedia~\cite{chen-yih-2020-open}. 
Despite the rapid progress in this area~\cite{chen-etal-2017-reading,karpukhin-etal-2020-dense,NEURIPS2020_6b493230}, the systems have primarily been evaluated in English, yet open-retrieval QA in non-English languages has been understudied~\cite{longpre2021mkqa,asai2021xor}. 
Moreover, due to the task complexity, cross-lingual open-retrieval QA has unique challenges such as multi-step inference (retrieval and answer selection) and cross-lingual pattern matching~\cite{lewis-etal-2020-mlqa,97cross-languageinformation}, whereas other multilingual NLP tasks have their inputs specified at once (e.g. natural language inference) and typically only need to perform inference on one language at a time.

In this work, we introduce the MIA 2022 shared task on cross-lingual open-retrieval QA, which tests open-retrieval QA systems across typologically diverse languages. 
Compared to previous efforts on multilingual open-retrieval QA~ \cite{10.5555/1813809.1813852,forner-etal-2010-evaluating}, this shared task covers a wider set of languages (i.e., 16 topologically diverse languages) and orders of magnitude more passages in retrieval targets (i.e., 40 million passages in total), and constitutes the first shared task for massive-scale cross-lingual open-retrieval QA.
Four teams submitted systems, three of which significantly improve the baseline system based on a state-of-the-art multilingual open-retrieval QA system~\cite{asai2021cora}.

Our analysis reveals that the system performance varies across languages even when the questions are parallel (as in one of our two settings), and several findings from the submitted systems shed light on the importance on entity-enhanced representations, leveraging more passages and data augmentation for future research in multilingual knowledge-intensive NLP. 
{
Our analysis suggests that (i) it is still challenging to retrieve passages cross-lingually, (ii) generating answers in the target language whose script differs from the script of evidence document is nontrivial, (iii) and potential answer overlaps in existing datasets may overestimate models' performance. 
}

We formally introduce our task in \Sref{sec:task}, followed by data collection process for 16 languages in \Sref{sec:data}. 
We then introduce our baseline systems in \Sref{sec:baseline} and the submitted systems. \Sref{sec:shared_task} presents our meta analysis of the systems performances, and we conclude by suggesting future improvements in this area. 

%% file: sections/task.tex
\section{Task Descriptions}
\label{sec:task}
We first formulate cross-lingual open-retrieval QA and introduce metrics used to evaluate systems' performance. We then present two submission tracks: constrained and unconstrained tracks.  

\subsection{Task Formulation}
Cross-lingual open-retrieval QA is a challenging multilingual NLP task, where given questions written in a user's preferred language, a system needs to find evidence from large-scale document collections written in many different languages. 
The final answer needs to be in the user’s preferred language which is indicated by their question, as in real-world applications.
We follow the general definition of \citet{asai2021cora}, where a system can retrieve evidence from documents in {\it any} languages, not limiting the retrieval target to certain languages as in \citet{10.5555/1813809.1813852}. 
For instance, a system needs to answer in Arabic to an Arabic question, but it can use evidence passages written in any language included in a large-document corpus such as English, German, Japanese and so on. 
In real-world applications, the issues of information asymmetry and information scarcity~\cite{roy2022information,blasi-etal-2022-systematic,asai2021xor,joshi-etal-2020-state} arise in many languages, hence the need to source answer contents from other languages---yet we often do not know \textit{a priori} in which language the evidence can be found to answer a question. 

\subsection{Evaluation Metrics}
\label{sec:metrics}

Systems are evaluated using automatic metrics: token-level F1 and exact match (EM).
Although EM is often used as the primary evaluation metric for English, the risk of surface-level mismatching~\cite{pmlr-v133-min21a} can be more pervasive in cross-lingual settings. 
Therefore, we use F1 as the primary metric and rank systems using the F1 scores. Evaluation is conducted using language-specific tokenization and evaluation scripts provided in the \miast{} repository.\footnote{For non-spacing languages (i.e., Japanese, Khmer, and Chinese), we use off-the-shelf tokenizers including Mecab, khmernltk and jieba to tokenize both predictions and ground-truth answers.}
We use data from \xorqa{} and \mkqa{} (detailed in \Sref{sec:data}), and due to different characteristics these datasets have, we macro-average scores per language set on each dataset, and then macro-average those scores to produce an F1 score for \xorqa{} and an F1 score for \mkqa{} to compute the final scores for ranking. 

\subsection{Tracks}
For the shared task, we defined two tracks based on the resource used to train systems: \textit{constrained} and \textit{unconstrained} settings. Systems trained only on the official training data qualify for the constrained track, while systems trained with additional data sources participate in the unconstrained track. 

\paragraph{Constrained Track.}
\label{sec:constrained-task}

To qualify as a constrained track submission, participants are required to use the official training corpus, which consists of examples pooled from \xorqa{} and Natural Questions~\cite{kwiatkowski2019natural}. See more data collection details in \Sref{sec:data}.
No other QA data may be used for training. 
We allow participants to use off-the-shelf tools for linguistic annotations (e.g. POS taggers, syntactic parsers), as well as any publicly available unlabeled data and models derived from these (e.g. word vectors, pre-trained language models).
In the constrained setup, participants may not use external blackbox APIs such as Google Search API and Google Translate API for inference, as those models are often trained on additional data, but they are permitted to use them for offline data augmentation or training.

\paragraph{Unconstrained track.}
\label{sec:unconstrained-task}

Any model submissions using APIs or training data beyond the scope of the constrained track are considered for the \textit{unconstrained} setting.
Participants are required to report the details of their additional resources used for training, for transparency. For instance, a submission might use publicly available QA datasets, such as CMRC 2018~\cite{cui-etal-2019-span} and FQuAD~\cite{dhoffschmidt-etal-2020-fquad}, to create larger-scale training data. 

%% file: sections/data.tex
\section{Shared Task Data}
\label{sec:data}
The \miast{} data is derived from two large-scale multilingual evaluation sets: \xorqa{}~\citep{asai2021xor} and \mkqa{}~\citep{longpre2021mkqa}.
{We first discuss the source datasets, and then discuss how the target languages are selected, and how the data is split into training and evaluation sets.}
\Tref{tab:language_family} shows the included languages, their language groups, the size of training, development and test data, and the number of Wikipedia passages available in each language.

\begin{table*}[t!]
\small
    \centering
    \begin{tabular}{l ll  rrr  r }
\toprule
 & \multicolumn{2}{c}{Language Family} &  \multicolumn{3}{c}{\# of examples} & \# Wiki. passages  \\\cmidrule(l){2-3}\cmidrule(l){4-6}
Language & Family & Branch & Train & Development & Test & \\\midrule
Arabic (ar) & Afro-Asiatic  & Semitic & 18,402 & 3,145 & 5,590 & 1,304,828\\
Bengali (bn) & Indo-European & Indo-Iranian & 5,007 &  2,248 & 5,203& 179,936 \\
English (en) & Indo-European & Germanic & 76,635 &  1,758 & 5,000 & 18,003,200 \\
Spanish (es) & Indo-European & Italic & 0 &  1,758 & 5,000 & 5,738,484\\
Finnish (fi) & Uralic &  Finnic & 9,762 & 2,732 & 1,368 & 886,595\\
Japanese (ja)  & Japonic & Japonic & 7,815 & 2,451 & 6,056 & 5,116,905\\
Khmer (km) & Austroasiatic & Khmer & 0 &  1,758 & 5,000 & 63,037 \\
Korean (ko) &Koreanic & Han & 4,319 & 2,231 &  6,048 & 638,864 \\
Malay (ms) & Austronesian & Malayo-Poly. & 0 &  1,758 & 5,000 & 397,396\\
Russian (ru) & Indo-European & Balto-Slavic & 9,290 & 2,776 & 6,910 & 4,545,635\\
Swedish (sv) & Indo-Europea & Germanic & 0 &  1,758 & 5,000 & 4,525,695\\
Chinese (zh) & Sino-Tibetan & Sinitic & 0 & 1,758 & 5,000 & 3,394,943 \\
Telugu (te) & Dravidian & South-Central & 6,759 & 2,322 & 6,873 &  274,230 \\\midrule
\textbf{Surprise Languages} & & & & \\
Tagalog (tl) & Austronesian  & Malayo-Poly. & 0 & 0 & 350  & -- \\ 
Tamil (ta) & Dravidian & Southern  &  0 & 0 & 350 & -- \\
 \bottomrule
 \end{tabular}
    \caption{List of the languages, their families and amount of data available in the MIA shared task data. The last two languages are surprise languages hidden from the participants. 
    }
    \label{tab:language_family}
\end{table*}

\subsection{Source Datasets}
\paragraph{\xorqa{}{} ~\citep{asai2021xor}} is a cross-lingual open-retrieval QA dataset covering 7 languages built upon TyDi QA~\cite{clark-etal-2020-tydi}. 
\citet{asai2021xor} collect answers for questions in TyDi QA that are \textit{unanswerable} using the same-language Wikipedia.
As the questions are inherited from TyDi QA, they are written by native speakers to better reflect their own interests and linguistic phenomena, and they are not parallel across languages. We use data for the XOR-full setting, where some questions can be answered based on the target language's Wikipedia (monolingual) while others require evidence only presented in English Wikipedia (cross-lingual). We use all of the 7 languages covered by \xorqa{}: Arabic (ar), Bengali (bn), Finnish (fi), Japanese (ja), Korean (ko), Russian (ru), Telugu (te). 
\paragraph{\mkqa{} ~\cite{longpre2021mkqa}} comprises the largest set of languages and dialects (26) for open-retrieval QA, spanning 14 language families.
There are 10k question and answer pairs per language.
The questions are human-translated from English Natural Questions~\cite{kwiatkowski2019natural} and the answers are re-annotated for higher quality -- chosen independently of any web pages or document corpora.
From \mkqa{}, we sample the 6,758 parallel examples which are answerable.
We select 12 of the 26 languages to lower the computational barrier:
Arabic (ar), English (en), Spanish (es), Finnish (fi), Japanese (ja), Khmer (km), Korean (ko), Malay (ms), Russian (ru), Swedish (sv), Turkish (tr), and traditional Chinese (zh-cn).

\subsection{Language Selection} 

We select a subset of languages from each resource (i) to cover a wide range of languages and typological features with a sufficient scale, and (ii) to compare participating model performance between questions that are translated from English and ones that are naturally generated by native speakers.
The natively-written questions from XOR-TyDi QA allow measuring systems' quality on questions that are likely to serve information need expressed by speakers of each language, whereas the human-translated questions of MKQA allow measuring the performance on the target script and language, holding constant the question content.
For this reason, we include 5 languages present in both \xorqa{} and \mkqa{} to compare the gap between cultural and linguistic model generalization: Arabic, Finnish, Japanese, Korean, and Russian.

\paragraph{Surprise languages.}
In addition, we newly annotated data in Tagalog (tl) and Tamil (ta){, where little work studies open-retrieval QA~\cite{liu-etal-2019-xqa}. }
For each language, we sample 350 \mkqa{} English examples, where the answer entities have an Wikipedia article in the target language.
The 350 questions are all translated using Gengo's human translation,\footnote{\url{https://gengo.com/}} but the answers are automatically translated using Wikidata.
This annotation results in 350 well-formed examples in Tagalog (tl) and Tamil (ta). 
Surprise languages are released two weeks before the system submission deadline to test systems' ability to perform zero-shot transfer~\cite{pmlr-v119-hu20b} to unseen languages that are substantially different from the languages they are trained on. 
{Except for one system, all of the submissions directly apply their systems to the new languages without any training or adding new target languages' Wikipedia.  }

\subsection{Data Statistics}
\Tref{tab:language_family} presents the list of the languages and statistics of the train, development and test set data in each target language.  

\paragraph{Training data.} 
Our training data consists of Natural Questions~\cite{kwiatkowski2019natural} for English and \xorqa{} for the other languages in the shared task.\footnote{See the training data linked at \url{https://github.com/mia-workshop/MIA-Shared-Task-2022\#training-data}}
In the constrained track (\Sref{sec:unconstrained-task}) only this data source is permitted for providing QA supervision, though other tools are permissible for data augmentation.

\paragraph{Evaluation data.} Our evaluation sets span 16 languages: 7 from \xorqa{} and 12 from \mkqa{} with an overlap of five languages and two surprise languages newly annotated for this shared task following MKQA annotation schema. 
We found that the original \xorqa{} validation and test splits have different proportions of the in-language and cross-lingual questions, resulting in large performance gaps between dev and test subsets as reported by \citet{asai2021cora}. 
We re-split \xorqa{} so that the validation and test sets have similar ratios of the two question types {of in-language and cross-lingual questions. }
{In-language questions are answerable from Wikipedia in the question's language, and are often easier to answer while the other category requires cross-lingual retrieval between the target language and English, and are more challenging. }
Further, we add aliases that can be retrieved via the Wikimedia API to the gold answers, following MKQA, thereby avoiding penalizing models for generating correct answers with surface-level differences. 
For \mkqa{} we split the answerable examples into a validation set of 1,758 questions and a test set of 5,000 question.
We add the newly annotated data for the surprise languages (Tamil and Tagalog) to the test set only.

\subsection{Limitations} 

\paragraph{False negatives in evaluations.}
First, because the original source questions and answers are from TyDi QA or Natural Questions, their answers are annotated based on a single Wikipedia article in English or the question language. 
{
MKQA answers are re-labeled by English speakers without any Wikipedia or web corpus, but small portion of the answers can be geographically incorrect for that regions of the languages the data is translated into (e.g., when the first harry potter movie was released?). }
As we generalize the task setting to {\it cross-lingual open retrieval}, there are inconsistent contents across articles in different languages leading to many possible answers.
However, because we only have one answer, this can penalize correct answers~\cite{palta2022investigating}.
It is a common issue that open-retrieval QA datasets do not comprehensively cover all valid answers~
\cite{pmlr-v133-min21a,asai-choi-2021-challenges}, and this can be more prevalent in multilingual settings due to transliteration of entities or diverse ways to express numeric in some languages~\cite{al-onaizan-knight-2002-translating}.

\paragraph{English American-centric biases.}
Second, the \mkqa{} questions as well as the new data annotated for this shared task are translated from English.
This annotation scheme enables us to scale up to many typologically diverse languages, but the resulting questions are likely to be Western- or specifically American-centric, rather than reflecting native speakers' interests and unique linguistic phenomena~\cite{clark-etal-2020-tydi}.
We try to reduce such English-centric bias by only using the questions whose answer entities are also included in Tamil or Tagalog Wikipedia, though this constrains the distribution to simple factoid questions.
We also found that in some languages, MKQA answers have high overlap with their English counterparts. 

%% file: sections/baselines.tex
\section{Baseline Models}
\label{sec:baseline}
We use a state-of-the-art open-retrieval QA model as our baseline. We open source the code, trained checkpoints, training data, and intermediate/final prediction results.\footnote{\url{https://github.com/mia-workshop/MIA-Shared-Task-2022}}
\subsection{Modeling}
Our baseline model is based on CORA~\cite{asai2021cora}, which has two components: mDPR for document retrieval and mGEN for answer generation. 
Both mDPR and mGEN are based on multilingual pretrained models to process data written in many different languages without relying on external translation modules. 

Given a question $q^L$ written in a language $L$, mDPR $\mathcal{R}$ retrieves top $N$ passages: $\mathbf{P} = p_1, \ldots, p_N = \mathcal{R}(q^L)$.  
mDPR includes all of the target languages' Wikipedias as its retrieval target, except for the two surprise languages.  
mGEN $\mathcal{G}$ takes as input $q$ and $\mathbf{P}$ and generates an answer $a^L$ in the target language: $a^L = \mathcal{G}(q, \mathbf{P})$. 
mDPR is a multilingual extension of DPR~\cite{karpukhin-etal-2020-dense}, which employs a dual-encoder architecture based on BERT~\cite{devlin-etal-2019-bert} and retrieves top passages based on the dot-product similarities between encoded representations. 
During training, mDPR optimizes the loss function as the negative log likelihood of the positive passages. 
mGEN simply concatenates the question and a set of top $K$ passages, and the fine-tuned multilingual encoder-decoder model generates a final answer in the target language. 
Unlike some prior work in English conducting end-to-end training of the retriever and reader \cite{lewis2020retrieval,guu2020realm}, we train mDPR and mGEN independently. 
Note that during mGEN training, we use the passages retrieved by the trained mDPR, as in \citet{izacard2021distilling}.

\subsection{Training and Hyperparameters}
We use the official training data for training. We also leverage the long answer annotations in the Natural Questions dataset and the gold paragraph annotations of \xorqa{} to create mDPR training data, released at the shared task repository.\footnote{\url{https://github.com/mia-workshop/MIA-Shared-Task-2022\#training-data}}
After training mDPR, we run it on the shared task training data questions to obtain top passages, and then use those retrieved passages to train the mGEN model: mGEN is trained to generate the gold answer given an input query and top retrieved passages. 

mDPR uses multilingual BERT-base uncased ~\cite{devlin-etal-2019-bert}, and mGEN is fine-tuned from mT5-base~\cite{xue-etal-2021-mt5}. For mDPR, we use the same hyperparameters as in DPR~\cite{karpukhin-etal-2020-dense}, and train it for 30 epochs, and take the last checkpoint. 
For mGEN, we follow \citet{asai2021cora} hyperparameters.

\subsection{Pre-processing Knowledge Corpus.}
Following DPR and mDPR, we split each article into 100-token chunks based on whitespace. 
For non-spacing languages (e.g., Japanese, Thai), we tokenize the articles using off-the-shelf tokenizers (i.e., MeCab for Japanese\footnote{\url{https://taku910.github.io/mecab/}.} and Thai NLP for Thai\footnote{\url{https://github.com/PyThaiNLP/pythainlp}.}). 
We exclude passages with less than 20 tokens. 
Total numbers of passages for each language are listed in \Tref{tab:language_family}. 

%% file: sections/submissions.tex
\section{Shared Task Submissions}
\label{sec:shared_task}
{Four teams submitted their final systems to our EvalAI~\cite{yadav2019evalai} leaderboard,\footnote{\url{https://eval.ai/web/challenges/challenge-page/1638/leaderboard}} three of which significantly outperformed the original baseline described in Section~\ref{sec:baseline}.}
We summarize the submitted systems here and refer readers to their system description paper for details. 
All of the systems are constrained submissions, and we did not receive any submission to unconstrained track. 

\begin{table*}[t!]
\centering
\resizebox{\textwidth}{!}{%
    \begin{tabular}{l| cc c| ccccccc }
\toprule
System & \multicolumn{3}{c}{Macro F1} &  \multicolumn{7}{c}{Language F1}   \\
 & Total & XOR & MKQA & Arabic & Bengali & Finnish & Japanese & Korean & Russian & Telugu \\\midrule
 (a) Texttron & {\bf 32.02} & {\bf 45.50} & 18.54 &  {\bf 56.37} &{\bf 42.43}&{\bf 43.13}& {\bf 44.71} & {\bf 34.37} & {\bf 47.79} &{\bf 49.72} \\
(b) mLUKE-FID & 31.61 &   40.93 & 22.29 & 45.33&30.48&41.01&43.45&31.21&42.62&42.40 \\
(c) CMUmQA & 31.53&40.20  & {\bf 22.87} & 55.06&30.56&41.25&42.44&28.76&42.56&40.75 \\
(d) ZusammenQA & 27.00 &37.95 & 16.04 & 49.66&33.99&39.54&39.72&25.59&40.98&36.16\\
\rowcolor[gray]{0.90} (e) Baseline & 27.55 & 37.95 & 17.14 &  51.66&31.96&38.68&40.89&25.35&39.87&37.26\\
 \bottomrule
 \end{tabular}
 }
    \caption{Final results on the \xorqa subsets of the MIA 2022 shared task. 
    The grayed entry indicates the baseline system. 
    }
    \label{tab:main_results_xorqa}
\end{table*}

\begin{table*}[t!]
\centering
\resizebox{\textwidth}{!}{%
    \begin{tabular}{l| cccccccccccccc }
\toprule
sys & \multicolumn{14}{c}{Language F1}   \\
  &   ar & en & es & fi & ko & ma & ja & km & ru & sv & tr & zh & tm & ta  \\\midrule
(a) & 13.62&33.24&28.98&25.26&13.07&29.04&23.11&3.96&20.11&29.75&{\bf 28.15}&11.30 & 0.00 & 0.00 \\
(b)  & 12.67&39.63&30.85&25.22&12.81&29.09&20.49&2.36&18.82&29.62&26.16&{\bf 22.60} &{\bf 20.75} &20.95 \\
(c) & {\bf 13.94} &{\bf 42.58}&{\bf 32.11}&{\bf 26.75} &{\bf 14.59} &{\bf 31.13} &{\bf 22.72}& {\bf 8.71} &{\bf 22.36} & {\bf 31.48} &26.59&18.00&2.74&{\bf 26.42} \\
(d)&  8.73&35.32&25.54&20.42&6.78&24.10&14.27&6.06&12.01&25.97&20.27&13.95&0.00&11.14 \\
\rowcolor[gray]{0.90}(e)  & 9.52&36.34&27.23&22.70&7.68&25.11&15.89&6.00&14.60&26.69&21.66&13.78&0.00&12.78 \\
 \bottomrule
 \end{tabular}
 }
    \caption{Final results on the MKQA subsets of the MIA 2022 shared task. 
    The grayed entry indicates the baseline.
    }
    \label{tab:main_results_mkqa}
\end{table*}

\paragraph{mLUKE+FiD.}
\citet{team_utah} adapt the retrieve-then-read baseline system with several improvements, including (a) using an mLUKE encoder \citep{ri2021mluke} for dense retrieval, (b) combining sparse and dense retrieval, (c) using a fusion-in-decoder reader \citep{izacard2021leveraging}, and (d) leveraging Wikipedia links to augment the training data with additional target language labels.

For retrieval, \citet{team_utah} use the $2019/02/01$ Wikipedia snapshot as their document corpora, matching the baseline.
They include the Wikipedia snapshots for Tamil and Tagalog to evaluate on the surprise languages.
Their sparse retriever searches the monolingual corpora only, while their dense retriever searches all corpora.

\paragraph{CMUmQA.}
\citet{cmumqa} build a four-stage pipeline for a retrieve-then-read approach, based on the CORA open-retrieval system \cite{asai2021cora} that searches evidence documents in any language for target questions (many-to-many QA; \citealp{asai2021cora}), without relying on translation.
They first apply an mBERT-based DPR retrieval model, followed by a reranker \cite{qu-etal-2021-rocketqa} with XLM-RoBERTA \cite{conneau-etal-2020-unsupervised}.
While it is computationally intractable to use for retrieval, the reranker has the advantage of encoding a question and a passage together, rather than independently.
An mT5-based fusion-in-decoder is then applied to generate an answer.
As the final step of their pipeline, Wikidata is used to translate English entities in the answer into the target language, if any.

\paragraph{ZusammenQA.}
\citet{ZusammenQA} follow the retrieve-then-read system, but with the expansion of several components, along with training methods and data augmentation.
Their retriever ensembles supervised models (mDPR and mDPR with a MixCSE loss; \citealp{wang2022sncse}) along with unsupervised sparse (Oracle BM-25) and unsupervised dense models (DISTIL, LaBSE, MiniLM, MPNet).

The reader system is based on mGEN, but with domain adaptation by continued masked language modeling on the document corpora, to better adapt to Wikipedia and the target languages.
The training data is augmented using \citet{dugan2022feasibility} that generates question-answer pairs from raw document corpora and translates them into multiple languages.

\paragraph{Texttron.} 
This submission also follows the retrieve-then-read structure: the retrieval model performs dense passage retrieval with XLM-RoBERTa Large \cite{conneau-etal-2020-unsupervised}, and the reading model uses mt5 large.
The retrieval text is split into paragraphs (as opposed to 100-word text segments) extracted by the WikiExtractor package.
The retrieval model is trained on a combination of three types of custom training data: target-to-target (both the query and retrieved paragraphs are in the target language), target-to-English (the query is in the target language and the retrieval paragraphs are in English), and English-to-English (both the query and retrieved paragraphs are in English). 
These data are created based on BM25 retrieval and query translation.

Texttron also used multiple stages of training and negative sample mining to tune their final dense retriever with hard negatives: a combination of BM25 and examples from the previous iteration of retrieval that had low token overlap with the gold answers.
No system description was available.

%% file: sections/results.tex
\section{Main Results}
\label{sec:res_analysis}
Tables \ref{tab:main_results_xorqa} and \ref{tab:main_results_mkqa} show final results on \xorqa and \mkqa{} subsets, respectively. 

\paragraph{Macro performance.}
Texttron, mLUKE + mFiD, and CMUmQA significantly improve the baseline performance. 
Among the constraint submissions, Texttron yields the best performance. 
While several systems achieve higher than 40 average F1 on \xorqa, only two systems achieve higher than 20 average F1 on MKQA, demonstrating how difficult it is to build a system that performs well in many languages without language-specific supervision.  
Texttron significantly outperforms other baselines on \xorqa while CMUmQA shows the best MKQA performance among the submitted systems. 

\paragraph{Language-wise performance.}
The performance varies across different languages.
Among \xorqa, all of the systems struggle in Korean and Bengali, while in Arabic, Japanese and Russian, they generally show relatively high F1 scores. 

On MKQA, where all of the questions are parallel, the performance still significantly differs across languages. 
Almost all of the systems report lower than 10 F1 in Khmer and Tamil, which are less represented in existing pretraining corpora~\cite{xue-etal-2021-mt5} and use their own script systems---with the notable exception of mLUKE + FiD, which achieves 20.8 F1 on Tamil.
mLUKE + FiD achieves substantially better performance than other systems in Tamil. 
This is partially because they also include the Tamil Wikipedia passages for passage retrieval, while other systems, including the baseline, do not.
As discussed in ~\citet{asai2021cora}, all systems show lower scores in the languages that are distant from English and use non-Latin scripts (e.g., Cyrillic for Russian, Hangul for Korean).

%% file: sections/analysis.tex
\section{Analysis}
{
We provide further analysis on the submitted systems. In \Sref{sec:summary_systems} we provide a brief summary of the findings from the submitted system descriptions. \Sref{sec:perf_comparison} provides performance comparison over answer-type, and answer overlap with English or training data. 
We then analyze the degree of answer agreements among the submitted systems to understand which questions remain challenging in ~\Sref{sec:prediction_agreement}. 
We further conduct manual error analysis in five languages in \Sref{sec:error_analysis}. 
}

\subsection{Summary of Findings}
\label{sec:summary_systems}
In this section, we highlight several effective techniques from the submitted systems.
Overall, a surprisingly wide range of complementary, and potentially additive, methods all reported strong benefits, including: (i) larger and longer pre-trained models for retrieving and reading, (ii) a reranking step with fusion-in-decoder multi-passage cross-encodings, (iii) iterative dense retrieval tuning with progressively harder negative example mining, (iv) using entity-aware retrieval encodings, (v) combining dense and sparse retrievers, (vi) data augmentation, and (vii) leveraging Wikidata answer post-processing for language localization. We discuss some of these below.

These findings highlight various techniques migrating the performances in English retrieval systems. 
And most of all, they emphasize that cross-lingual retrieval still poses the major bottleneck to the end-to-end task, while large multilingual fusion-in-decoder reader systems can operate well when given sufficient evidence.
These findings suggest multilingual retrieval is the most important avenue for future research, especially on questions not easily answered by English Wikipedia. Moreover, retrieving evidence cross-lingually is keys for other knowledge intensive NLP tasks such as fact verification~\cite{thorne-etal-2018-fever} and knowledge-grounded dialogues~\cite{dinan2018wizard} beyond open-retrieval QA. 

\paragraph{Entity representations.}
Using entity-aware representations for the passage retriever's encoders gives a large performance improvement; As shown in analysis by Team Utah~\cite{team_utah}, replacing mBERT encoders in DPR with mLUKE improves by 1.22 F1 on XOR macro-average and 1.85 MKQA macro F1. 
We hypothesize that the mLUKE may capture better cross-lingual entity alignment than mBERT as it leverages inter-language links in Wikipedia during pretraining. 
This sheds light on the potential effectiveness of multilingual entity contextualized representations for cross-lingual passage representations, which is an under-explored direction.  

\paragraph{Combining dense and sparse retrievers \& hard negatives.}
Texttron and Team Utah combine both BM25 and mDPR, while ZusammenQA explore a diverse set of unsupervised and supervised retrieval approaches including BM25 and LaBSE~\cite{feng-etal-2022-language}. 
Team Utah shows that combining BM25 with mDPR helps, while ZusammenQA shows that only using BM25 gives significantly lower scores than the original baseline~\cite{ZusammenQA}, as BM25 does not have cross-lingual phrase matching capabilities.
Texttron iteratively trained their dense retriever, mining increasingly hard negative examples using BM25 and query translation, filtered using simple heuristics.

\paragraph{Fusion-in-Decoder and passage reranking.}
Team Utah and CMUmQA demonstrate that Fusion-in-Decoder architectures outperform simply concatenating passages as in mGEN (Fusion-in-Encoder). While Fusion-in-Encoder simply concatenates retrieved passages in a retrieved order, Fusion-in-Decoder encodes each of the retrieved passages independently and then concatenate them. This may help the model to pay more attentions to the passages that are ranked lower by the retriever but indeed provides evidence to answer. 
Recent work in open domain QA also demonstrates that the Fusion-in-Decoder architecture is more competitive than prior systems that simply concatenate passages~\cite{fajcik-etal-2021-r2-d2,asai2021evidentiality}.

Team Utah show increasing the number of passages improves performance, while CMUmQA show that cross-encoder reranking is particularly beneficial for Fusion-in-Decoder. 

\paragraph{Data augmentation.}
ZusammenQA introduces data augmentation using Google Translate to translate the training data into target languages. {\bf AUG-QA} translates question-answer pairs into target languages, while {\bf AUG-QAP} translates question, answer and the original training data passages into the target languages. They found that the AUG-QAP and AUG-QA both improve performance from their direct counterpart without data augmentation. 

\paragraph{Wikipedia answer localization.}
CMUmQA and others used Wikidata entity maps to localize answers to the correct target script following~\citet{longpre2021mkqa}.
This process was particularly effective for localizing short answers into a target language from English due to the overwhelming English bias of retrieval and generative systems finetuned on English. As a result, CMUmQA obtains the best MKQA performance among the submitted systems. 

\subsection{Performance Comparison}
\label{sec:perf_comparison}
In this section, we group questions based on several factors (e.g., answer types) and compare the models' performance across different sub-groups. 

\paragraph{Answer types.}
MKQA provides answer categories for each question. We analyze the per-category model performance to understand what types of questions remain challenging. 
The original MKQA source data except for the unanswerable subsets has the following answer type distributions: Entity (42\%),  Date (12\%), Number (5\%), Number with Unit (4\%), Short Phrase (3\%), Boolean (yes, no; 1\%), Unanswerable (14\%), and Long Answers (13\%).
The Unanswerable and Long Answers categories are excluded from the MIA 2022 shared task evaluation data.  

\begin{table}
\small
    \centering
    \begin{tabular}{l| rrrr}
\toprule
& en & es & ja & zh \\\midrule
Number with units & 7.77 & 3.56 & 1.94 & 3.88 \\
Entity & \textbf{58.18} & \textbf{53.19} & 34.42 & 15.75 \\
Number & 27.07 & 29.83 & 21.27 & 25.70 \\
Date & 28.14 & 28.49 & 6.10 & 11.37 \\
Short phrases & 8.60 & 7.81 & 5.08 & 5.08 \\
Binary & 32.99 & 31.96 & \textbf{79.38} & \textbf{75.25} \\
 \bottomrule
 \end{tabular}
    \caption{The percentage of the exact match per answer types in English (en), Spanish (es), Japanese (ja) and Chinese (zh).
    }
    \label{tab:per_category}
\end{table}

{
We present the percentage of the questions where any of the submitted system predictions match the annotated gold answers in English, Spanish, Japanese and Chinese in \Tref{tab:per_category}. 
In all of the languages, the systems show relatively higher exact matching rate in Entity types questions except for Chinese and Japanese. In those languages, many of the entity names are written in their own script systems (e.g., Chinese characters, katakana), which is challenging to be generated from the evidence passages written in other languages; it is known to be challenging to translate an entity name from one language to another using different script systems~\cite{wang-etal-2017-sogou}. 
In English and Spanish, the systems show significantly higher accuracy on entity and date than in Japanese or Chinese, while the systems struggle in Boolean questions. 
\xorqa Japanese subset shows higher percentage of boolean questions than other subsets, which potentially helps the systems in Japanese and Chinese MKQA boolean questions. 
All of the systems show significantly lower performance in short phrase questions, indicating the difficulty of generating phrase length answers beyond simple factoid questions with entity or date answers. 
}

\begin{figure*}
\centering
\begin{subfigure}[t]{.52\linewidth}
  \centering
  \includegraphics[width=0.95\textwidth,keepaspectratio]{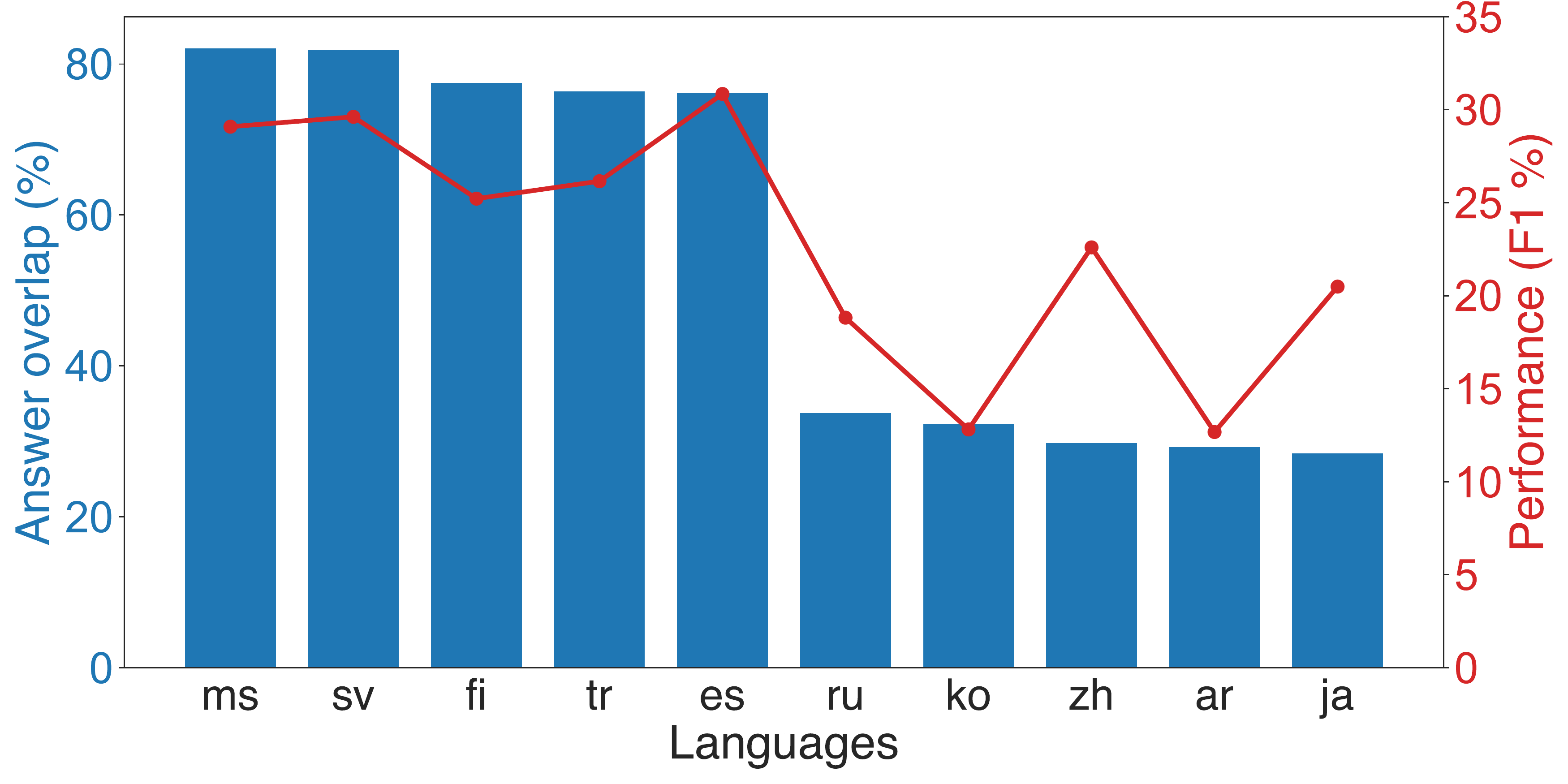}
   \captionsetup{width=0.95\textwidth}
  \caption{MKQA performance vs. answer overlap with English answers.}
  \label{fig:performance_answer_overlap}
\end{subfigure}%
\begin{subfigure}[t]{.47\linewidth}
  \centering
  \includegraphics[width=0.95\textwidth,keepaspectratio]{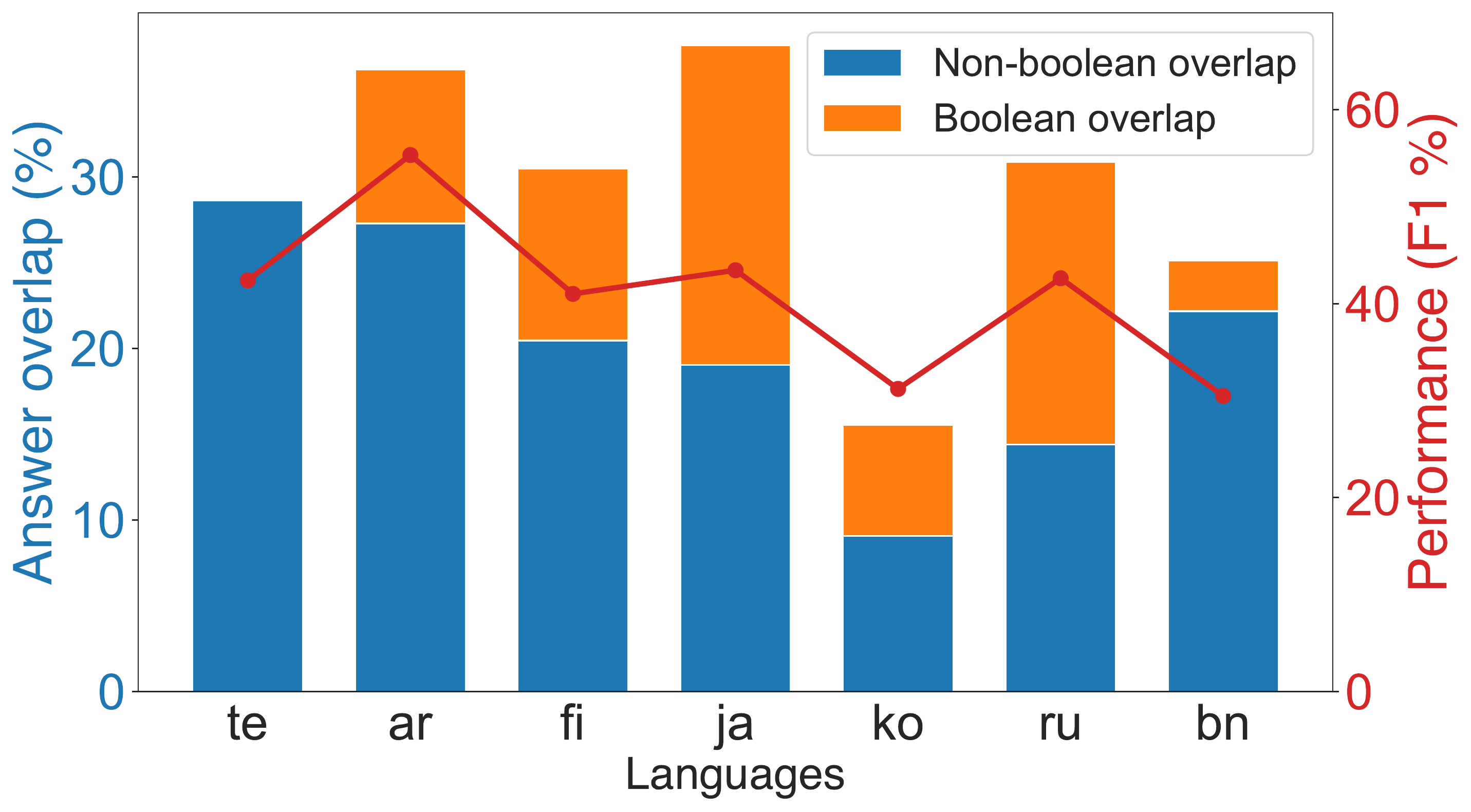}
   \captionsetup{width=0.95\textwidth}
  \caption{\xorqa~performance vs. answer overlap between train and test sets.}
  \label{fig:train_test_overlap}
\end{subfigure}%
\caption{Performance vs. answer overlap between train and test sets.
}
\end{figure*}

\paragraph{Answer overlaps with English.}
We analyze performances across languages by examining the relationship between the final performance and the number of the questions whose answers are the same as English answers.
\Fref{fig:performance_answer_overlap} shows mLUKE + FiD, which is the second best system in our shared task, and answer overlap with the English subsets for each MKQA language except for Khmer and two surprise languages. 
We observe a clear correlation between the answer overlap and final performance among those languages. The model performs well on the languages where many answers are the same as English answers. 
Finnish, on the other hand, shows relatively lower performance compared to other languages with high answer overlap (i.e., Malay, Swedish, Spanish). 
Among the languages with low answer overlap, on the Japanese and Chinese sets, the system shows relatively high F1 scores compared to the other languages with lower than 40\% overlap (i.e., Russian, Korean, Arabic). {This is likely 
because Chinese and Japanese show higher accuracy on Boolean type questions than other languages as discussed above. 
}

\paragraph{Answer overlap with training data.}
Prior work shows that the high overlap between train and test data can result in the overestimated performance of the systems~\cite{lewis-etal-2021-question}. 
In \xorqa, the questions are annotated by native speakers of the target languages, so the percentage of the train-test overlap can vary across languages. 
{
We calculate the percentage of the answers for the test data questions that also appear as gold answers in \xorqa~training data. 
We then check whether the degree of the answer overlap between the train and test sets correlate with the final \xorqa~test performance.}


\begin{figure*}
\centering
\begin{subfigure}[t]{.32\linewidth}
  \centering
  \includegraphics[width=0.95\textwidth,keepaspectratio]{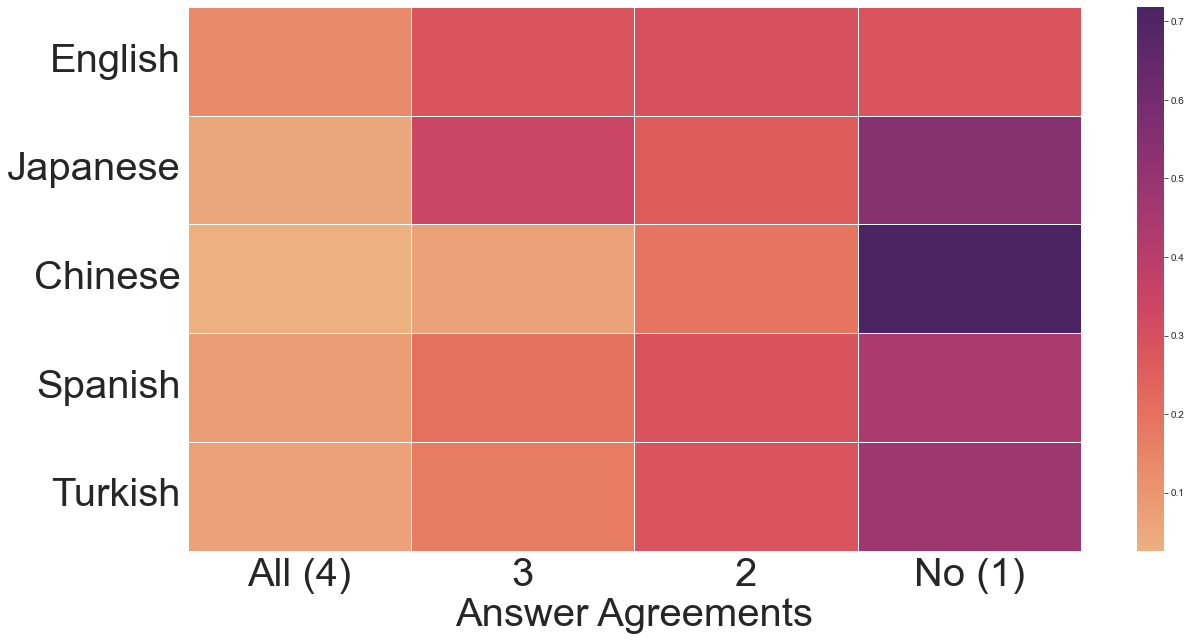}
   \captionsetup{width=0.95\textwidth}
  \caption{MKQA Answer agreement. }
  \label{fig:answer_overlap}
\end{subfigure}%
\begin{subfigure}[t]{.335\linewidth}
  \centering
  \includegraphics[width=\textwidth,keepaspectratio]{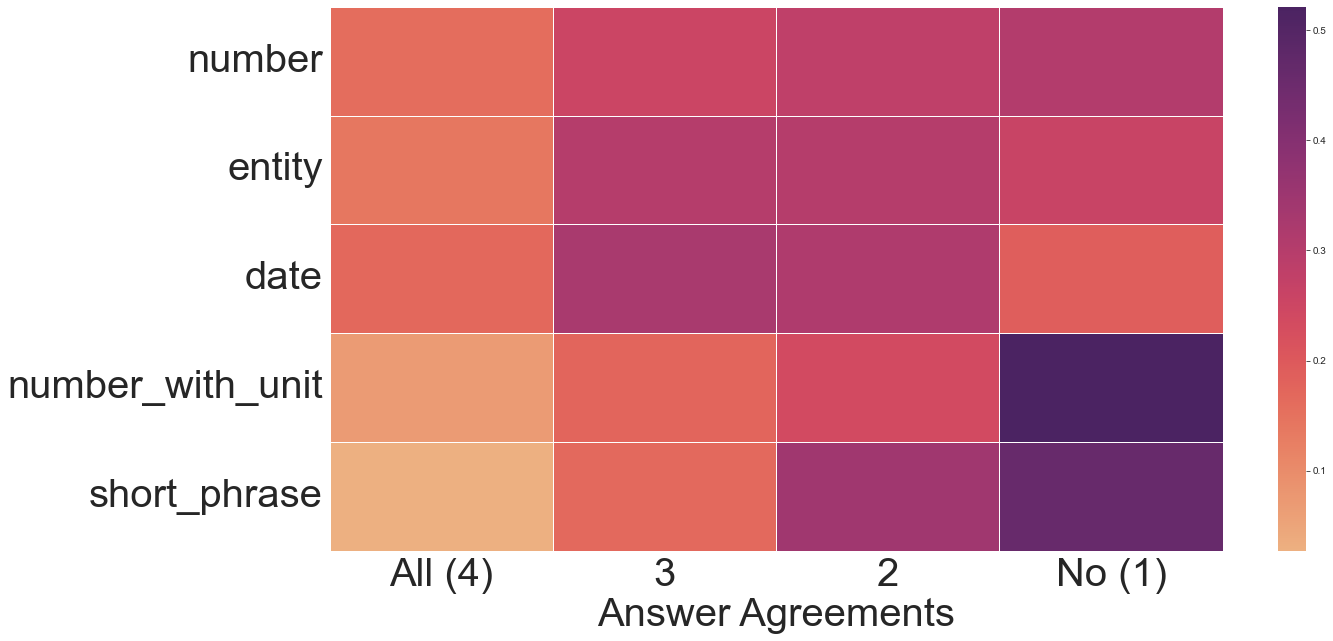}
   \captionsetup{width=0.95\textwidth}
  \caption{Per-category agreement (En).}
  \label{fig:category_agreement_en}
\end{subfigure}%
\begin{subfigure}[t]{.335\linewidth}
  \centering
  \includegraphics[width=\textwidth,keepaspectratio]{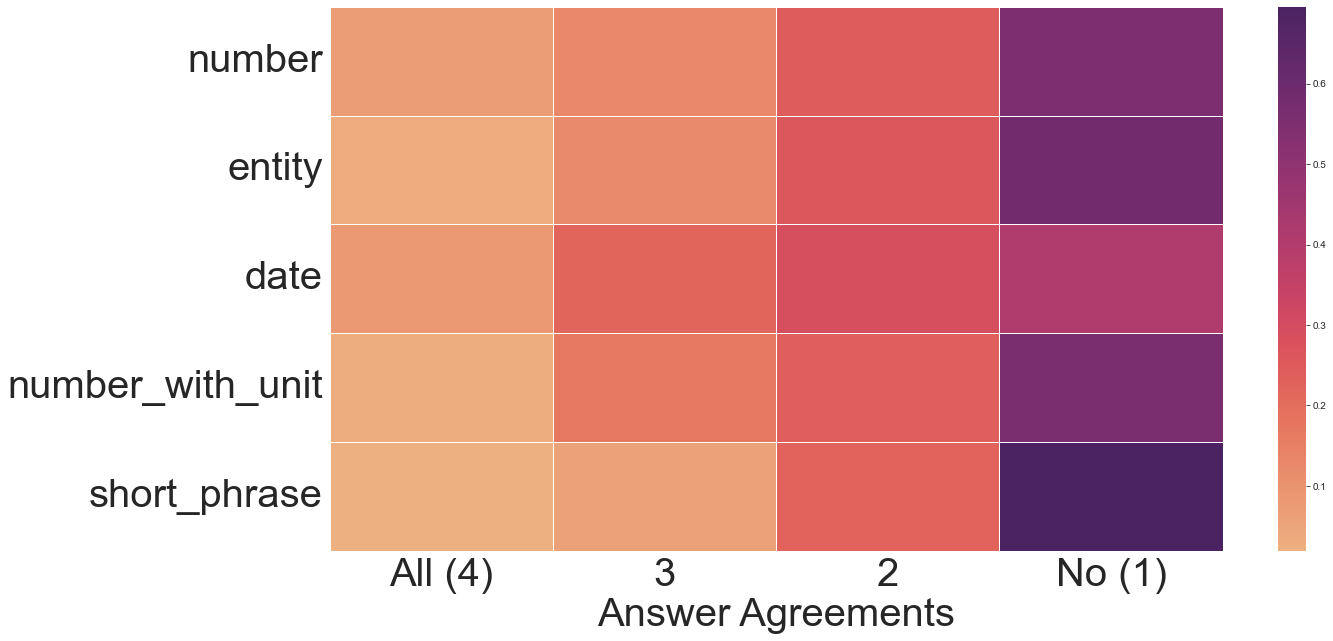}
   \captionsetup{width=0.95\textwidth}
  \caption{Per-category agreement (Ja).}
  \label{fig:category_agreement_ja}
\end{subfigure}%
\caption{Answer agreements of the four submitted systems.
}
\end{figure*}

\Fref{fig:train_test_overlap} shows the performance and train-test overlap percentage. Although we can see the percentage of overlap between train and test data varies across languages, it is not particularly correlated with the final performance.
For instance, Bengali actually shows relatively high overlap between train and test data (over 25\% answer overlap), but the performance is much lower than Telugu, whose answer overlap ratio is close to that of Bengali. 
We also found that the percentage of the Boolean questions (yes, no) significantly differs across languages: in Japanese, around 10\% of the questions are Boolean questions, while in Telugu, almost no questions are Boolean. The original TyDi QA data is annotated by different groups of annotators for each language, and thus such question distributions can differ~\cite{clark-etal-2020-tydi}. 

\paragraph{\xorqa~vs. \mkqa{}. }
Arabic, Japanese, Korean, and Finnish are included both in \mkqa{} and \xorqa, but their performance on the two subsets significantly differ; In general, the \xorqa F1 scores are much higher than \mkqa{} (e.g., Japanese: 44.71 vs. 23.11). 
We hypothesize that this happens because we do not have training data for \mkqa{} and all \mkqa{} questions tend to require cross-lingual retrieval as the questions are translated from English and answers are American-centric.
In contrast, half of the questions in \xorqa~are from TyDi QA, and the answers are grounded to their own languages' Wikipedia. 
Cross-lingual retrieval is generally more challenging than monolingual retrieval~\cite{zhang-etal-2021-mr}. In addition, all of the \xorqa cross-lingual questions are labeled ``unanswerable'' in TyDi QA, and can be more difficult to answer than its monolingual counterparts. 

\begin{table*}
\small
    \centering
    \begin{tabular}{l|cc|cc|cc|cc|cc|cc|cc|cc}
\toprule
 & \multicolumn{2}{c}{Avg.} &  \multicolumn{2}{c}{Arabic} & \multicolumn{2}{c}{Bengali} & \multicolumn{2}{c}{Finnish} & \multicolumn{2}{c}{Japanese} & \multicolumn{2}{c}{Korean} & \multicolumn{2}{c}{Russian} & \multicolumn{2}{c}{Telugu} \\
Sys. & cl &  m & cl & m &   cl & m &  cl & m  &  cl & m &  cl & m  &  cl & m &  cl & m \\\midrule
(a)  & 27.2 & 58.8 & 28.3 & 64.8 & 29.4 & 63.3 & 30.3 & 51.3 & 29.5 & 55.9 & 22.2 & 53.2& 24.9 &55.8  & 25.9 & 67.6  \\
(b) &  21.4 & 54.2 & 22.2 & 65.2  & 21.7 &44.6  & 24.1 &  51.7& 27.4 & 55.2 &  19.5 & 49.3& 19.1 &50.8 & 16.0 & 62.2 \\
(c)  & 20.3 & 52.5 & 21.5 & 64.6 & 20.5  & 42.3 & 25.5 & 50.9 & 26.3 & 53.1 & 17.6 & 44.9 & 16.0 &  51.2 & 14.7 & 60.3 \\
(d)  & 19.5 & 49.7 & 22.4 & 59.0 & 19.0 & 51.5 & 23.8  & 46.9 & 25.2 & 50.2 &  15.3 & 38.8 & 16.9 & 47.0 & 14.0 & 54.2  \\
 \bottomrule
 \end{tabular}
    \caption{Final results on MIA 2022 Shared Task \xorqa cross-lingual (``cl'') / monolingual subsets (``m''). Systems (a), (b), (c) and (d) are Texttron, mLUKE-FID, CMUmQA, and ZusammenQA, respectively. 
    }
    \label{tab:cross_mono_performance}
\end{table*}

{
To further test this hypothesis, we evaluate the submitted systems' performance on \xorqa's cross-lingual and monolingual subsets in \Tref{tab:cross_mono_performance}. 
We can clearly see that all of the baseline's performance deteriorates on the cross-lingual subsets, while they show high F1 scores across languages on the monolingual subsets. }


\subsection{Prediction Agreement}
\label{sec:prediction_agreement}
We analyze how often all of the systems agree on the same answers on the \mkqa{} test data in five languages. In particular, we compare all of the four system predictions on the English, Japanese, Chinese, Spanish and Turkish subsets of the \mkqa{} test data, and check the prediction agreements based on the number of the unique predictions among the union of the predictions. 
We can see that in English and Spanish, the agreement is high (e.g., in 40\% of the questions, all or three of the four systems agree on the same answers), while the agreement is lower in other languages, particularly in Japanese and Chinese. 

To understand the phenomena, we breakdown the prediction agreement statistics in English and Japanese into different answer categories. \Fref{fig:category_agreement_en} and \Fref{fig:category_agreement_ja} show per-category prediction agreements in English and Japanese, respectively. 
While in English, systems show high agreements in date, entity and number type questions, in Japanese, the agreement rate is lower across category, potentially because of their diverse formats of number and dates, as well as the transliteration of the entity names.

\subsection{Error Analysis}
\label{sec:error_analysis}

We conduct a set of error analysis in five languages (i.e., English, Japanese, Korean, Chinese and Telugu) on randomly sampled 30 questions, where none of the submission systems' predictions exactly match any of the ground truth answers. 

\paragraph{Error types.}

We classify the errors into following categories: (i) incorrect predictions, (ii) answers are semantically correct in different languages (incorrect languages), (iii) incorrect gold answers, (iv) semantically-equivalent predictions in the target language but are penalized because gold answers do not cover all of the potential gold answers (not comprehensive gold answers),  (v) questions are open-ended or ambiguous (e.g., entity ambiguity), (vi) questions' granularity is unclear (unclear question granularity; e.g., year v.s. month, kilometers v.s. meters), (vii) questions are highly subjective (e.g., who is the best singer ever), (viii) temporal or geographical dependency in questions.

The first two error types, (i) and (ii), reveal the limitations of models. 
The error type (iii) and (iv) are considered answer annotation errors~\cite{pmlr-v133-min21a,asai-choi-2021-challenges}. 
The last four error types (v), (vi), (vii) and (viii) requires some specifications or context~\cite{zhang-choi-2021-situatedqa,min-etal-2020-ambigqa}. 

\paragraph{Error analysis schema.}
We recruit native speakers of the five target languages and ask them to classify the errors into the aforementioned categories. 
We present the predictions of all of the systems as well as the intermediate retrieval results of the mLUKE + FiD.

\begin{table*}
\small
    \centering
    \begin{tabular}{l| cccccc}
\toprule
& English & Arabic & Japanese & Korean & Chinese  \\\midrule
(i) incorrect predictions & 12 & 9 & 23 & 16 & 12 \\
(ii) incorrect languages & 0 & 2 &  3 & 0 & 2 \\
(iii) incorrect gold answers  & 2 & 4 &5  &1  & 0 \\
(iv) not comprehensive gold answers & 10 & 1  & 7 & 5 & 6 \\
(v)  ambiguous question & 3 & 7 &  6 & 15 & 5 \\
(vi) unclear question granularity & 3 & 2  & 1 & 2 & 0 \\ 
(vii) subjective question & 0 & 0  &  0&0  &0 \\
(viii) temporal or geographical dependency in questions & 4 &  4 &  1 & 4 & 5 \\
 \bottomrule
 \end{tabular}
    \caption{Error analysis on sampled questions where all of the submissions unanimously fail to predict the correct answers. We show the percentage of the errors in each category.
    }
    \label{tab:error_analysis}
\end{table*}
\paragraph{Error analysis results.}
\Tref{tab:error_analysis} provides the error analysis result. 
Besides modeling errors, we found that the original annotations themselves exhibit some issues, which underestimates models' performance. Across languages,  annotators found non-negligible proportion of the errors happen as the original gold answers do not cover all of the possible answer aliases or the answer granularity is unclear. For instance, an English question asks ``what is the temperature at the center of earth'' and the gold answer is 6000 °C. 
Several systems answer in Fahrenheit or Kelvin, and got zero F1 score. 
Several questions are also temporal or geographical dependent such as ``who was the last person appointed to the u.s. supreme court'' or \begin{CJK}{UTF8}{min}クリミナル・マインドの新シーズンが公開されるのはいつか\end{CJK} (when is the next season of Criminal Minds will be released?).
Although situation-grounded QA has been recently studied~\cite{zhang-choi-2021-situatedqa}, there's little work that analyzes this phenomena in multilingual settings, where the particularly geographical dependence can be even more prevalent.   
Question ambiguity is also common in multilingual QA.





%% file: sections/conclusion.tex
\section{Conclusion and Discussions}
We have presented the MIA 2022 Shared Task on cross-lingual open-retrieval QA systems in 16 typologically diverse languages, many of which are unseen during training. 
Several submissions improved significantly over our baseline based on a state-of-the-art cross-lingual open-retrieval QA system and investigated a wide range of techniques. 
Those results shed light on the effectiveness of several techniques in this challenging task, such as entity-enhanced representations, sparse-dense retrieval, and better interactions between passages. 
We further conducted detailed performance analysis on different subsets of the datasets, such as languages, answer types, the necessity of cross-lingual retrieval as well as detailed error analysis.
We also suggest several bottlenecks in the area.  